\documentclass[twocolumn]{article} % Or use 'IEEEtran' for IEEE format, etc.

\usepackage{multicol}  % For two-column support
\usepackage[letterpaper, top=0.6in, bottom=0.5in, inner=0.75in, outer=0.75in, headsep=0.3in, footskip=0.3in]{geometry}

\usepackage{titlesec}
\titleformat{\section}{\large\bfseries}{\thesection}{1em}{}
\titleformat{\subsection}{\normalsize\bfseries}{\thesubsection}{1em}{}

\usepackage{graphicx}         % For including images
\usepackage{amsmath, amssymb} % For math symbols and equations (even if you don't use much, good practice)
\usepackage{hyperref}          % For hyperlinks (URLs, references)
\usepackage{natbib}          % For bibliography management (BibTeX style)
\usepackage{geometry}         % For adjusting margins if needed
\usepackage{listings}
\usepackage{pgfplots}
\usepackage{amssymb} % For \checkmark
\usepackage{pifont}  % For \xmark
\usepgfplotslibrary{fillbetween}
\usepackage{algorithm}
\usepackage{algorithmic}
\usepackage{amsmath} 
\usepackage{appendix}

\title{\textbf{ReZero: Enhancing LLM search ability by trying one-more-time}}

\author{
    Alan Dao (Gia Tuan Dao)\textsuperscript{1}, Thinh Le\textsuperscript{1}\\
    Menlo Research \\
    \texttt{alan@menlo.ai, thinh@menlo.ai} \\
    \textsuperscript{1}Equal contribution.
}

\date{\today} % Or specify a date, or use \date{} for no date

\begin{document}

\maketitle
\begin{abstract}
 % Prevents indenting the abstract paragraph
\noindent Retrieval-Augmented Generation (RAG) improves Large Language Model (LLM) performance on knowledge-intensive tasks but depends heavily on initial search query quality. Current methods, often using Reinforcement Learning (RL), typically focus on query formulation or reasoning over results, without explicitly encouraging persistence after a failed search. We introduce ReZero (Retry-Zero), a novel RL framework that directly rewards the act of retrying a search query following an initial unsuccessful attempt. This incentivizes the LLM to explore alternative queries rather than prematurely halting. ReZero demonstrates significant improvement, achieving 46.88\% accuracy compared to a 25\% baseline. By rewarding persistence, ReZero enhances LLM robustness in complex information-seeking scenarios where initial queries may prove insufficient.
\end{abstract}

% --- Include Sections ---
\section{Introduction}
\label{sec:introduction}

Large Language Models (LLMs) have demonstrated impressive capabilities in understanding and generating human language, yet they often struggle with tasks requiring access to up-to-date, specific, or proprietary knowledge not captured during their pre-training phase \cite{10.1145/3637528.3671470,jeong-etal-2024-adaptive}. Retrieval-Augmented Generation (RAG) has emerged as a dominant paradigm to address this limitation, equipping LLMs with the ability to query external knowledge sources, typically search engines or vector databases, to ground their responses in relevant information \cite{kim2024suresummarizingretrievalsusing}.

However, the effectiveness of RAG systems hinges critically on the quality of the interaction between the LLM and the retrieval system. Complex questions often necessitate multi-step reasoning, where information gathered in one step informs subsequent queries or reasoning processes \cite{trivedi-etal-2023-interleaving,yao2023react}. Even with advanced techniques, the initial query formulated by an LLM might be suboptimal, ambiguous, or fail to retrieve the necessary information on the first attempt. Existing approaches often focus on refining the reasoning process over retrieved documents \cite{madaan2023selfrefine}. For instance, methods like ReARTeR \cite{sun2025rearterretrievalaugmentedreasoningtrustworthy} utilize Process Reward Models (PRMs) and Process Explanation Models (PEMs) to score and refine intermediate reasoning steps within a RAG pipeline, aiming for trustworthy multi-step reasoning. Other approaches, such as DeepRetrieval \cite{jiang2025deepretrievalhackingrealsearch}, employ reinforcement learning (RL) to directly optimize the query generation process itself, using retrieval metrics like recall or NDCG as rewards to train the LLM to formulate more effective queries through trial and error.

\begin{figure}
    \centering
    \includegraphics[width=1\linewidth]{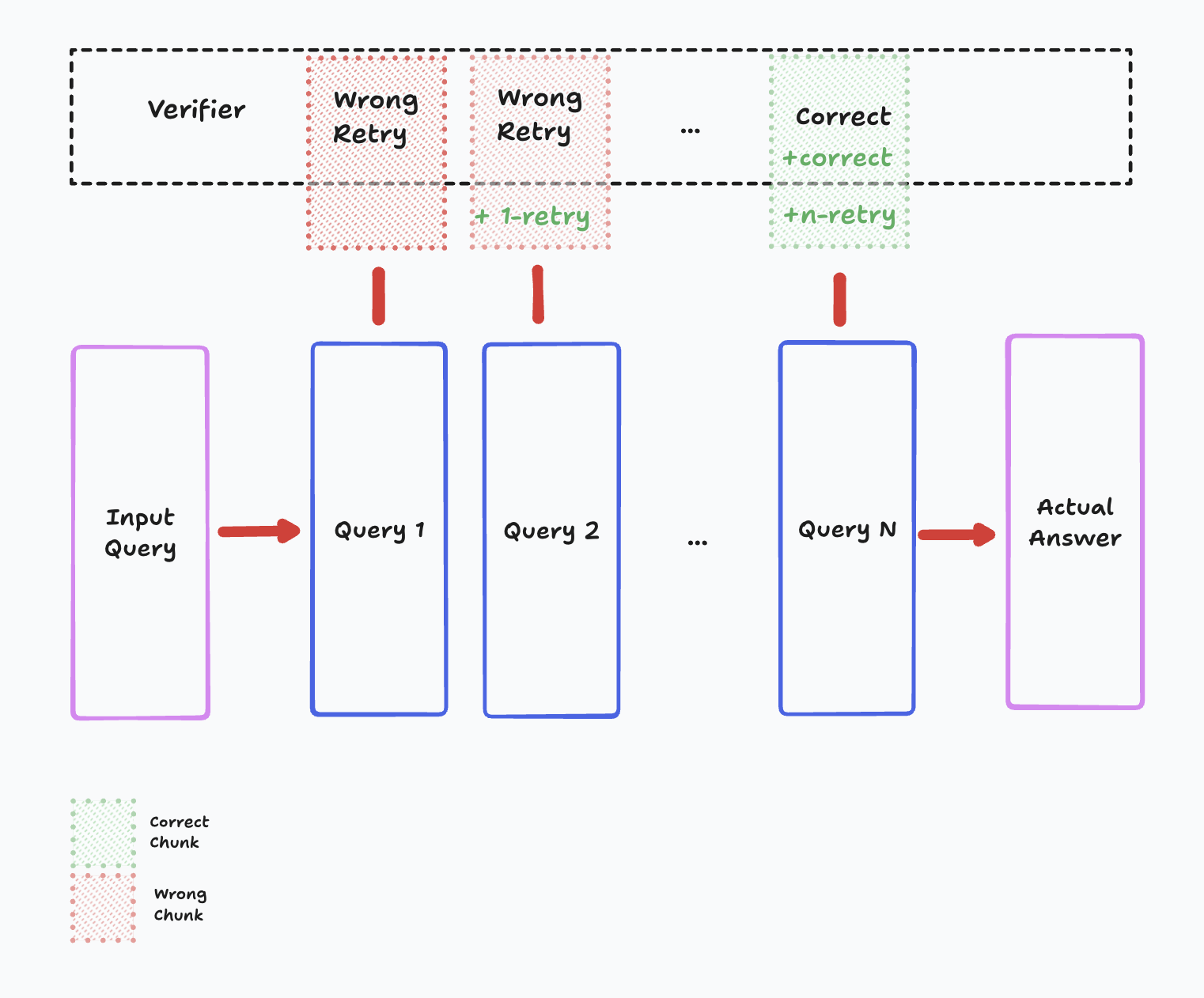}
    \caption{Receives a reward signal for retrying after failure.}
    \label{fig:concept-demo}
\end{figure}

While these methods enhance reasoning quality and query effectiveness, they primarily reward the successful outcome of a reasoning step or a retrieval action. They often implicitly assume that a single, well-formed query or reasoning step is the goal. However, they may not sufficiently incentivize the model to persist when an initial search attempt yields unsatisfactory results. The inherent difficulty in formulating a perfect query from the outset, especially for complex or exploratory information needs, suggests that the ability to recognize inadequacy and retry the search process could be a valuable, yet underexplored, capability.

To address this gap, we introduce ReZero (Retry-Zero), a novel framework designed to improve LLM search ability by explicitly incentivizing the model to "try one more time." Our core idea is simple: we augment the standard reinforcement learning process for RAG with a specific reward component that encourages the model to retry its search query if the initial attempt is deemed insufficient. Unlike approaches focusing solely on the correctness of the final answer or the relevance of retrieved chunks, ReZero directly rewards the act of retrying the search interaction. This is based on the intuition that persistence-trying again, potentially with a reformulated query based on the initial failure or intermediate reasoning-can significantly improve the chances of eventually finding the necessary information, echoing the real-world problem-solving strategy: "if at first you don't succeed, try, try again."

ReZero employs a reinforcement learning strategy where the LLM interacts with a search environment. The reward function is designed not only to reflect the quality of the final answer but also to provide a positive signal when the model executes a "retry" action following an initial search. This encourages the model to explore different querying strategies and learn when persistence is beneficial, rather than giving up or hallucinating after a single failed attempt.

The main contributions of this work are:

\begin{itemize}
    \item We propose ReZero, a novel RL-based framework that explicitly rewards the act of retrying search queries within a RAG system, fostering persistence in information seeking.
    \item We introduce a modified reward function that incentivizes the "retry" mechanism, encouraging exploration and potentially leading to better performance on complex, knowledge-intensive tasks where initial searches may fail.
    \item We position ReZero relative to existing work on RAG reasoning and query optimization, highlighting its unique focus on rewarding persistence rather than solely immediate outcomes.
\end{itemize}
\section{Related Work}
\label{sec:related_work}

Our work builds upon and extends several lines of research in enhancing LLMs, particularly in the context of Retrieval-Augmented Generation (RAG) and Reinforcement Learning (RL).

\subsection{Retrieval-Augmented Generation (RAG)} 

RAG systems \cite{10.1145/3637528.3671470,jeong-etal-2024-adaptive,xu2024recomp} have become a standard method for grounding LLM outputs in external knowledge, improving factuality and handling knowledge-intensive tasks. Early RAG approaches often involved a single retrieval step before generation. However, complex tasks frequently require multiple interactions with the knowledge source, leading to research in multi-step RAG \cite{trivedi-etal-2023-interleaving,yao2023react}. Methods like Self-Ask \cite{press-etal-2023-measuring} and IRCoT \cite{trivedi-etal-2023-interleaving} integrate Chain-of-Thought (CoT) reasoning with iterative retrieval, allowing the model to decompose questions and gather information incrementally. While these methods improve reasoning over retrieved context, they often rely on sophisticated prompting or assume the model inherently knows when and how to retrieve effectively. ReZero differs by focusing on the robustness of the retrieval interaction itself, specifically by encouraging retries when initial attempts fail, a dimension less explored in iterative RAG frameworks which often focus on sequential information gathering rather than correcting failed searches.

\subsection{Learning and Search for Reasoning in RAG} 

Recent work has explored using learning-based methods, particularly RL and process supervision, to enhance the reasoning capabilities of RAG systems, often following the "learning and search" principle [32].

\cite{sun2025rearterretrievalaugmentedreasoningtrustworthy} exemplifies this direction by focusing on trustworthy multi-step reasoning. It employs Process Reward Models (PRM) to score intermediate reasoning steps and Process Explanation Models (PEM) to provide critiques for refinement, using techniques like MCTS and preference optimization (KTO) \cite{10.5555/3692070.3692574,pang2024iterative}. ReARTeR aims to improve the quality and reliability of each reasoning step. While it uses rewards and refinement, its focus is on the correctness and trustworthiness of the reasoning process given the retrieved information. ReZero, in contrast, focuses on the search interaction itself, specifically incentivizing the model to retry the search if the initial attempt fails or seems inadequate, adding a layer of persistence before or during the reasoning-over-documents phase.

\cite{jiang2025deepretrievalhackingrealsearch} tackles the problem from the query generation perspective. It uses RL (PPO) to train LLMs to generate or rewrite queries that maximize retrieval performance (e.g., Recall, NDCG) on various retrieval systems, including real-world search engines. DeepRetrieval learns effective query formulation through trial-and-error, optimizing for the outcome of the search. ReZero complements this by adding a reward for the process of retrying. While DeepRetrieval optimizes the quality of a single (potentially refined) query attempt for maximal success, ReZero encourages the model to make multiple attempts if necessary, rewarding the persistence mechanism directly.

\subsection{Reinforcement Learning for LLM Alignment and Reasoning}

RL has become a cornerstone for aligning LLMs with human preferences (RLHF) \cite{NEURIPS2022_b1efde53} and enhancing specific capabilities like reasoning \cite{deepseekai2025deepseekr1incentivizingreasoningcapability} and tool use \cite{schick2023toolformer}. Methods like ReFT \cite{wu2024reft} use RL (e.g., PPO) to fine-tune LLMs for reasoning tasks based on outcome or process rewards. The concept of using rewards to shape LLM behavior is central to ReZero. However, while prior work has used RL to optimize reasoning steps [21, 41] or query generation \cite{jiang2025deepretrievalhackingrealsearch}, ReZero introduces a novel application of RL by incorporating a reward signal specifically tied to the retry action in a search context. This aligns with the broader goal of using RL to encourage desirable behaviors, but targets the specific behavior of persistence in information retrieval.

The idea of iterative improvement is also present in self-correction or self-refinement methods \cite{huang2024large,madaan2023selfrefine}, where LLMs critique and revise their own generated outputs. These methods typically focus on refining the generated text (e.g., reasoning steps, final answers) based on internal checks or external feedback (like verifier scores). ReZero differs in its focus: instead of refining the LLM's generated output, it encourages retrying the interaction with the external search tool, addressing potential failures at the information-gathering stage itself.

In summary, ReZero occupies a unique space by leveraging RL to explicitly incentivize persistence in the search process within RAG. While related to work on improving RAG reasoning (ReARTeR) and query optimization (DeepRetrieval), its core novelty lies in rewarding the "retry" mechanism, aiming to make LLMs more robust and effective information seekers, especially when faced with complex or initially ambiguous queries.
\section{Methodology}
\label{sec:methodology}

\subsection{Overview}
\label{subsec:overview}
ReZero is a reinforcement learning (RL) framework designed to enhance the search capabilities of large language models (LLMs) in retrieval-augmented generation (RAG) systems. Inspired by recent advancements in RL for reasoning tasks \cite{deepseekai2025deepseekr1incentivizingreasoningcapability} and motivated by findings suggesting RL fosters better generalization compared to supervised fine-tuning, ReZero utilizes Group Relative Policy Optimization (GRPO) \cite{shao2024deepseekmathpushinglimitsmathematical} to explicitly incentivize persistence—rewarding the model for retrying search queries when initial attempts fail. 

\subsection{Reinforcement Learning Framework}
\label{subsec:rl_framework}
ReZero operates within a search environment where the LLM interacts with an external retrieval system. We employ the Group Relative Policy Optimization (GRPO) algorithm, noted for its effectiveness in training LLMs for reasoning tasks without requiring a separate critic model. The RL loop involves standard components:
\begin{itemize}
    \item \textbf{State:} The current conversation history, including the user's prompt, the LLM's previous responses (potentially including \texttt{<search>} and \texttt{<information>} tags), and retrieved information.
    \item \textbf{Action:} The LLM's generation, which could be an internal thought process (\texttt{<think>}), a search query (\texttt{<search>}), a final answer (\texttt{<answer>}), or, critically, the decision to issue another search query after a previous one.
    \item \textbf{Reward:} A scalar signal derived from evaluating the LLM's outputs against predefined criteria using the reward functions described below. These functions collectively act as a self-teacher.
    \item \textbf{Policy:} The LLM's strategy for generating actions, fine-tuned using GRPO to maximize cumulative reward.
\end{itemize}
\subsection{Reward Functions}
\label{subsec:reward_functions}
ReZero employs multiple reward functions to provide the training signal for GRPO. These functions evaluate different aspects of the LLM's generation:
\begin{enumerate}
    \item \texttt{reward\_correctness}: Evaluates the final answer's accuracy against a ground-truth, checks response structure validity, and outputs a binary reward. The binary reward is determined by the model itself, acting as a self-judge (LLM-as-a-Judge) \cite{gu2025surveyllmasajudge} using the base model's capabilities.
    \item \texttt{reward\_format}: Ensures adherence to the required conversational format and tag usage (e.g., tag sequence, valid markup), outputting a binary reward.
    \item \textbf{\texttt{reward\_retry}}: This reward function encourages the model to persist when initial search attempts do not yield sufficient information. It assigns a positive reward for each subsequent \texttt{<search>} query issued after the first one within a single generation sequence (i.e., for retries). The magnitude of the reward could potentially diminish with each additional retry to encourage efficiency. Crucially, this reward is conditional on task completion, it is only awarded if the model's final generated output in the sequence includes the complete \texttt{<answer>...</answer>} tags. If the sequence concludes without a well-formed answer enclosed in these tags, the \texttt{reward\_retry} component contributes zero to the total reward for that trajectory, regardless of how many retries were performed. This mechanism prevents the model from learning to accumulate reward simply by retrying repeatedly without ever successfully generating a final answer.
    \item \texttt{reward\_em\_chunk}: Verifies if the correct information chunk was retrieved by comparing the content in \texttt{<information>} tags against a ground-truth chunk using exact matching, outputting a binary reward.
    \item \texttt{reward\_search\_strategy}: Evaluates the quality of the search process by checking adherence to a desired conversational flow: initiating a broad \texttt{<search>}, analyzing retrieved \texttt{<information>} (verified by specific keywords within \texttt{<think>} tags), executing subsequent refined \texttt{<search>} queries based on this analysis, and finally synthesizing an \texttt{<answer>} grounded in the analyzed information. Outputs a graded score (0.0-1.0) based on the successful execution of these sequential phases.
    \item \texttt{reward\_search\_diversity}: Assesses the variety within the sequence of <search> queries used during generation. It rewards distinct query concepts and semantic dissimilarity between queries (measured using normalized string comparison). Bonus rewards are allocated for the effective use of diverse search operators (e.g., site:, "", OR). Penalties are applied to discourage submitting exact duplicate or highly similar queries, promoting broader exploration. Outputs a graded score (0.0-1.0) rewarding unique queries and operator diversity, while penalizing repetition and high similarity.
\end{enumerate}
These functions collectively guide the policy towards correctness, format adherence, effective information gathering, and search persistence.
\subsection{Training Process}
\label{subsec:training}
The LLM is fine-tuned directly from a pre-trained base model using reinforcement learning, specifically employing the Group Relative Policy Optimization (GRPO) algorithm. The training operates within an interactive framework involving a verifier-in this case a search engine \cite{snell2025scaling}, drawing parallels to setups studied for improving model generalization through RL \cite{chu2025sftmemorizesrlgeneralizes}. The initial reference policy ($\pi_{\text{ref}}$) for GRPO is the base pre-trained model itself.

\begin{itemize}
    \item \textbf{Iterative Interaction Loop (Rollout):} The core training dynamic involves the LLM interacting with the search engine (verifier). As illustrated in Figure \ref{fig:concept-demo}, for a given input prompt, the LLM (starting from the base pre-trained policy) generates a response sequence. This sequence can include emitting a search query (\texttt{<search>}). The verifier processes this query and returns information chunks (\texttt{<information>}). The LLM receives these chunks and continues generating. Critically, the model might iteratively repeat this query-retrieval process within the same generation sequence if the initial results are deemed insufficient, before finally producing an answer (\texttt{<answer>}). This iterative generation forms one complete trajectory or rollout per input prompt.
    \item \textbf{Reward Calculation:} Upon completion of the entire generated sequence, the sequence is evaluated. The total reward for this trajectory is computed using the suite of reward functions described in Section \ref{subsec:reward_functions}(\texttt{reward\_correctness}, \texttt{reward\_format}, \texttt{reward\_retry}, \texttt{reward\_em\_chunk}). This aggregated reward reflects the overall quality and effectiveness of the generated sequence, including the persistence demonstrated through retries.
    \item \textbf{Policy Update (GRPO):} The calculated rewards for multiple sampled trajectories (generated using the current policy $\pi_{\theta}$) are fed into the GRPO algorithm. GRPO updates the LLM's parameters ($\theta$) by comparing the reward of each trajectory against the average reward of the group (batch), aiming to increase the probability of generating higher-reward sequences. A KL divergence term against the reference policy ($\pi_{\text{ref}}$, which is the initial base model) is typically used to stabilize training and prevent drastic deviations from the initial capabilities.
    \item \textbf{Noise Injection for Robustness:} To specifically strengthen the model's ability to generalize its retry strategy, we introduce noise during training at the vector database level. This simulates imperfect retrieval by randomly perturbing the relevance or quality of returned chunks for some queries. This encourages the LLM to learn robust retry mechanisms rather than overfitting to scenarios where the first search attempt always yields perfect results.
\end{itemize}

This process directly fine-tunes the base LLM using RL, teaching it not only to answer correctly but also to strategically and persistently use the search tool, even when facing retrieval imperfections, without an intermediate supervised fine-tuning stage.

\section{Experiments and Results}
\label{sec:experiments}
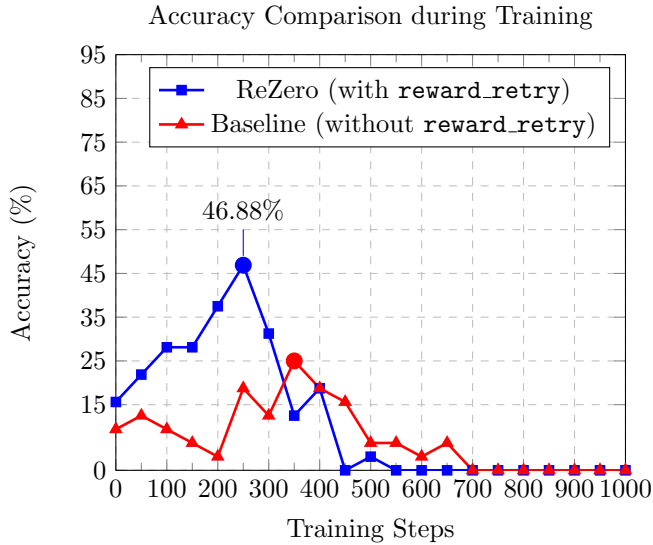
\begin{figure}[htbp]
\centering
\begin{tikzpicture}
\begin{axis}[
    title={Accuracy Comparison during Training},
    xlabel={Training Steps},
    ylabel={Accuracy (\%)},
    xmin=0, xmax=1000,
    ymin=0, ymax=95, % Adjusted ymax closer to peak
    xtick={0, 100, 200, 300, 400, 500, 600, 700, 800, 900, 1000},
    ytick={0, 15, 25, 35, 45, 55, 65, 75,85,95}, % Adjusted yticks
    legend pos=north east, % Specific outer position
    ymajorgrids=true,
    xmajorgrids=true,
    grid style=dashed,
    width=0.47\textwidth, % Adjusted width slightly for outer legend
    height=0.4\textwidth, % Adjusted height
    xticklabel style={/pgf/number format/1000 sep=},
    minor tick num=1,
]
% --- ReZero Plot ---
\addplot[
    color=blue,
    mark=square*,
    mark size=1.5pt,
    line width=1pt,
    ]
    coordinates {
    (0,15.62) (50,21.88) (100,28.12) (150,28.12) (200,37.50) (250,46.88) (300,31.25) (350,12.50) (400,18.75) (450,0.00) (500,3.12) (550,0.00) (600,0.00) (650,0.00) (700, 0.00) (750, 0.00) (800, 0.00) (850, 0.00) (900, 0.00) (950, 0.00) (1000,0.00)
    };
    \addlegendentry{ReZero (with \texttt{reward\_retry})}
% --- Highlight ReZero Peak ---
\addplot[
    only marks, % Only draw the marker
    mark=*, % Use a filled circle marker
    mark size=3pt, % Make it larger
    color=blue,
    forget plot % Don't add this to the legend
    ]
    coordinates {(250, 46.88)}; % The peak coordinate
% --- Add Text Label for ReZero Peak ---
\node[pin={[pin edge={blue, thin}, pin distance=10pt]90:{46.88\%}}] at (axis cs:250, 46.88) {}; % Simple label above
% Alternative label style (uncomment to use):
%\node[blue, anchor=south west, xshift=2pt, yshift=2pt, font=\small] at (axis cs:250, 46.88) {46.88\%};
% --- Baseline Plot ---
\addplot[
    color=red,
    mark=triangle*,
    mark size=2pt,
    line width=1pt,
    ]
    coordinates {
    (0,9.38) (50,12.50) (100,9.38) (150,6.25) (200,3.12) (250,18.75) (300,12.50) (350,25.00) (400,18.75) (450,15.62) (500,6.25) (550,6.25) (600,3.12) (650,6.25) (700,0.00) (750,0.00) (800,0.00) (850,0.00) (900,0.00) (950,0.00) (1000,0.00)
    };
    \addlegendentry{Baseline (without \texttt{reward\_retry})}
% --- Highlight Baseline Peak ---
\addplot[
    only marks,
    mark=*,
    mark size=3pt,
    color=red,
    forget plot
    ]
    coordinates {(350, 25.00)}; % The peak coordinate
% --- Add Text Label for Baseline Peak ---
% Alternative label style (uncomment to use):
%\node[red, anchor=north west, xshift=2pt, yshift=-2pt, font=\small] at (axis cs:350, 25.00) {25.00\%};
\end{axis}
\end{tikzpicture}
\caption{Comparison of evaluation accuracy between the ReZero model (incorporating the \texttt{reward\_retry} component) and the Baseline model (lacking the retry incentive) over 1000 training steps on the held-out Apollo 3 dataset chunks. Peak accuracies are highlighted.} % Added note to caption
\label{fig:accuracy_comparison_highlighted}
\end{figure}

Our experimental setup implements the ReZero framework detailed in Section 3. % Assuming Section 3 is Methodology
We fine-tuned a pre-trained large language model using Group Relative Policy Optimization (GRPO) \cite{shao2024deepseekmathpushinglimitsmathematical}, following the reinforcement learning approach described in the paper. The training process operates within an interactive environment involving a search engine acting as a verifier \cite{snell2025scaling}, without an intermediate supervised fine-tuning stage. The implementation utilized the \texttt{unsloth} training library and borrowed significantly from the codebase presented in \cite{autodidact2022}.
For our experiments, we employed the Apollo 3 mission dataset. This dataset was divided into 341 distinct data chunks, with 32 chunks specifically reserved for evaluating model performance. The training was executed on a single NVIDIA H200 GPU and ran for a total of 1000 steps, which corresponds to approximately 3 epochs over the training portion of the dataset (309 chunks).We choose \textbf{Llama3.2-3B-Insruct }as the base model for training \cite{grattafiori2024llama3herdmodels}.
To isolate and assess the impact of the proposed \texttt{reward\_retry} mechanism, we conducted experiments comparing two model configurations:
\begin{itemize}
    \item \textbf{Baseline:} This model was trained using three reward functions: \texttt{reward\_correctness}, \texttt{reward\_format}, and \texttt{reward\_emchunk}. It lacked the explicit incentive to retry search queries provided by the fourth reward function.
    \item \textbf{ReZero (with \texttt{reward\_retry}):} This model represents the full implementation of our proposed framework. It was trained using all four reward functions, crucially including the \texttt{reward\_retry} component designed to encourage persistence by rewarding subsequent search attempts within a single generation sequence, conditional on successful final answer generation.
\end{itemize}
Both models started from the same base pre-trained weights and underwent the same fundamental RL training procedure using GRPO, differing only in the inclusion of the \texttt{reward\_retry} signal. This controlled comparison allows for a direct evaluation of the contribution of rewarding the retry action.
Model performance was evaluated periodically on the held-out evaluation set (32 chunks) throughout the 1000 training steps. The primary metric reported is accuracy, defined as the percentage of correctly answered queries within the evaluation set.
The results, depicted in Figure~\ref{fig:accuracy_comparison_highlighted}, clearly indicate the effectiveness of the \texttt{reward\_retry} component. The ReZero model achieved a peak accuracy of \textbf{46.88\%} at 250 training steps. In contrast, the Baseline model, without the retry incentive, reached a maximum accuracy of only \textbf{25.00\%} at 350 steps. The ReZero model not only achieved a significantly higher peak performance but also demonstrated a faster initial learning rate compared to the baseline. Both models exhibited a decline in accuracy after reaching their peaks, potentially due to overfitting or instability in the later stages of RL training, with accuracy dropping to 0\% by step 700 for the Baseline and step 450 for the ReZero model (persisting at 0% through step 1000).

% --- End PGFPlots Figure ---

The substantial gap in peak accuracy (46.88\% vs 25.00\%) strongly suggests that explicitly rewarding the act of retrying search queries via the \texttt{reward\_retry} function significantly enhances the model's ability to effectively utilize the search tool and ultimately arrive at correct answers, particularly in scenarios where initial queries might be insufficient.
% Add bibliography commands if you use BibTeX/BibLaTeX
% \printbibliography

\section{Discussion}
\label{sec:discussion}

The experimental results presented in Section~\ref{sec:experiments} provide compelling evidence for the efficacy of the proposed ReZero framework, particularly the contribution of the \texttt{reward\_retry} component. This mechanism mirrors aspects of human information-seeking, where initial attempts often require refinement or alternative approaches. The most striking finding is the substantial performance gap observed between the ReZero model and the Baseline configuration, as depicted in Figure~\ref{fig:accuracy_comparison_highlighted}. The ReZero model, equipped with the incentive to retry search queries, achieved a peak accuracy of 46.88\%, nearly double the 25.00\% peak accuracy attained by the Baseline model which lacked this specific reward signal. This significant difference strongly validates our central hypothesis: explicitly rewarding the act of retrying a search query enhances the LLM's ability to navigate information retrieval challenges. Crucially, this reward was conditional on the successful generation of a final answer (as detailed in Section~\ref{sec:methodology}), ensuring it incentivized productive persistence rather than simply encouraging repeated, fruitless queries. The faster initial learning rate observed in the ReZero model further suggests that encouraging persistence accelerates the model's adaptation to effective search strategies. The use of noise injection during training likely amplified the scenarios where initial searches failed, potentially providing more effective learning signals for the \texttt{reward\_retry} mechanism compared to a cleaner retrieval environment, although the precise interaction warrants further study.

However, the performance trajectory also reveals a critical challenge and an area for future investigation. Both models, despite their differing peak performances, exhibited a notable decline in accuracy after reaching their respective zeniths (ReZero after 250 steps, Baseline after 350 steps), eventually collapsing to 0\% accuracy on the evaluation set. This phenomenon, while not uncommon in reinforcement learning \cite{chu2025sftmemorizesrlgeneralizes}, suggests that the current limitations may lie less in the core concept of rewarding retries and more in the specifics of the RL training process itself. Factors such as potential overfitting to the training data chunks, instability inherent in the GRPO algorithm over extended training, or suboptimal balancing of the different reward components could contribute to this degradation. The relatively short training duration (1000 steps, ~3 epochs) might not be sufficient to achieve stable convergence, or conversely, it might be pushing the model into unstable policy regions quickly. This points towards a need for further research into optimizing the RL training regime for sustained performance.

Furthermore, a significant limitation of this study is its reliance on a single dataset derived from the Apollo 3 mission. While this dataset provided a controlled environment for comparing the two model configurations, it represents a specific and relatively narrow domain. The types of questions, the nature of the information within the chunks, and the overall complexity might not be representative of the diverse scenarios LLMs encounter in real-world RAG applications. Therefore, the generalizability of the observed performance gains to broader knowledge domains or different types of information-seeking tasks remains an open question. The pronounced performance gap seen here might be amplified or diminished when applied to datasets with different characteristics.

Future research should prioritize addressing these limitations. Evaluating ReZero across a wider range of datasets spanning multiple domains and varying query complexities is essential to ascertain the robustness and general applicability of the approach. Concurrently, further investigation into stabilizing the RL training process is crucial. This could involve exploring alternative RL algorithms, refining hyperparameters, implementing advanced regularization, or dynamically adjusting reward components. Additionally, exploring variations of the \texttt{reward\_retry} function itself, such as incorporating diminishing returns or conditioning on retrieval quality improvement, could yield further benefits. It would also be valuable to conduct a qualitative analysis of the generated search queries to understand how the model adapts its strategy during retry attempts – whether it learns meaningful reformulations or simpler persistence patterns. Analyzing the practical trade-off between accuracy gains and the potential increase in latency or computational cost associated with additional search queries will also be important for real-world deployment. Finally, combining ReZero with orthogonal techniques, such as advanced query rewriting \cite{jiang2025deepretrievalhackingrealsearch} or sophisticated reasoning-over-retrieved-documents methods \cite{sun2025rearterretrievalaugmentedreasoningtrustworthy}, could lead to synergistic benefits.

\section{Conclusion}
\label{sec:conclusion}

This work addressed the challenge of enhancing the robustness of Retrieval-Augmented Generation (RAG) systems, particularly when initial search queries fail to retrieve the necessary information. We introduced ReZero (Retry-Zero), a novel reinforcement learning framework built upon Group Relative Policy Optimization (GRPO), designed to explicitly incentivize persistence in the search process. Unlike existing approaches that primarily focus on query formulation or reasoning over retrieved results, ReZero incorporates a specific reward component, \texttt{reward\_retry}, which encourages the Large Language Model (LLM) to attempt subsequent search queries following an unsuccessful initial attempt, conditional on successfully generating a final answer.

Our experiments, conducted on the Apollo 3 mission dataset, demonstrated the significant impact of this approach. The ReZero model, trained with the \texttt{reward\_retry} incentive, achieved a peak accuracy of 46.88\%, substantially outperforming the 25.00\% peak accuracy of a baseline model trained without this specific reward signal. This result strongly supports our hypothesis that explicitly rewarding the act of retrying enhances the LLM's ability to overcome initial search failures and improve task success rates.

Despite the promising results, we acknowledge limitations. The observed decline in performance after reaching peak accuracy highlights challenges related to the stability of the RL training process over extended periods, suggesting a need for further investigation into optimization techniques or regularization methods. Furthermore, the evaluation was confined to a single, specific domain (Apollo 3 mission data), limiting the current claims of generalizability across diverse knowledge areas and query types.

Future research should prioritize validating ReZero across a wider range of datasets and task complexities to ascertain its broader applicability. Stabilizing the RL training dynamics is crucial for achieving sustained performance. Exploring refinements to the \texttt{reward\_retry} function itself (e.g., diminishing returns, conditioning on retrieval improvement), conducting qualitative analyses of the learned retry strategies, investigating the latency and computational cost trade-offs, and exploring the integration of ReZero with complementary RAG techniques (like advanced query rewriting or reasoning methods) represent promising avenues for further enhancement.

In conclusion, ReZero offers a valuable contribution by demonstrating that directly rewarding persistence—the willingness to "try one more time"—can significantly improve the effectiveness of LLMs in complex information-seeking scenarios. This work highlights the potential of incorporating mechanisms that mirror human problem-solving strategies into the training of capable and robust AI systems operating within RAG frameworks.

\bibliographystyle{plainnat} % Choose your bibliography style (e.g., 'plainnat', 'abbrvnat', etc.)
\bibliography{bibliography} % 'bibliography' is the name of your .bib file (without the extension)
% Note down some papers here:

% --- Appendices (Optional) --- 
% --- Appendices (Optional) ---

% --- Appendix A: Detailed Experimental Setup ---
% \clearpage
% \clearpage
% \appendix
% \onecolumn
% \input{sections/appendices.tex}
% ... Appendix content ...

\end{document}